\documentclass[10pt,twocolumn,letterpaper]{article}
\pdfoutput=1

\usepackage{iccv}
\usepackage{times}
\usepackage{epsfig}
\usepackage{graphicx}
\usepackage{amsmath}
\usepackage{amssymb}

\usepackage[pagebackref=true,breaklinks=true,letterpaper=true,colorlinks,bookmarks=false]{hyperref}

\iccvfinalcopy 

\def\iccvPaperID{****} 

\ificcvfinal\fi
\begin{document}

\title{Semantic Video Segmentation by Gated Recurrent Flow Propagation}

\author{David Nilsson \qquad  Cristian Sminchisescu\\
Department of Mathematics, Faculty of Engineering, Lund University\\
{\tt\small \{david.nilsson, cristian.sminchisescu\}@math.lth.se}
}

\maketitle


\begin{abstract}

Semantic video segmentation is challenging due to the sheer amount of data that needs to be processed and labeled in order to construct accurate models. In this paper we present a deep, end-to-end trainable methodology to video segmentation that is capable of leveraging information present in unlabeled data in order to improve semantic estimates. Our model combines a convolutional architecture and a spatio-temporal transformer recurrent layer that are able to temporally propagate labeling information by means of optical flow, adaptively gated based on its locally estimated uncertainty. The flow, the recognition and the gated temporal propagation modules can be trained jointly, end-to-end. The temporal, gated recurrent flow propagation component of our model can be plugged into any static semantic segmentation architecture and turn it into a weakly supervised video processing one. Our extensive experiments in the challenging CityScapes and Camvid datasets, and based on multiple deep architectures, indicate that the resulting model can leverage unlabeled temporal frames, next to a labeled one, in order to improve both the video segmentation accuracy and the consistency of its temporal labeling, at no additional annotation cost and with little extra computation. 
\end{abstract}


\section{Introduction}

\begin{figure*}
\centering
\includegraphics[page=3,scale=0.45,trim=0cm 8.2cm 0cm 0cm]{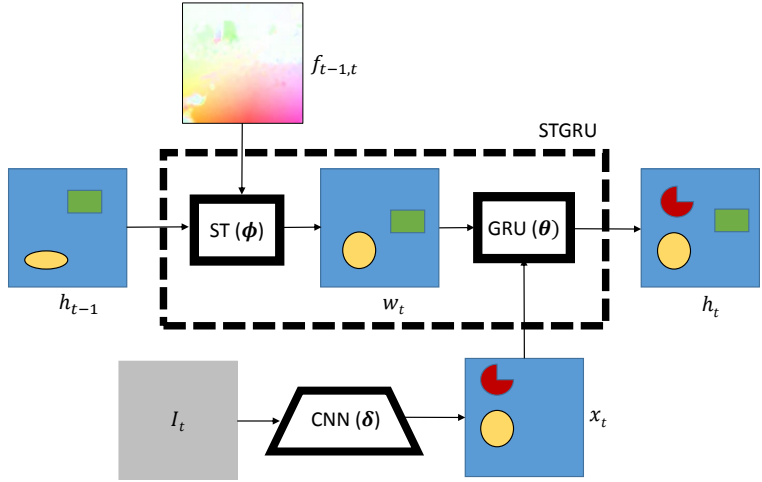}
\caption{Overview of our Spatio-Temporal Transformer Gated Recurrent Unit (STGRU), combining a Spatial Transformer Network (\S\ref{sec:stnet}) for optical flow warping with a Gated Recurrent Unit (\S\ref{sec:gru}) to adaptively propagate and fuse semantic segmentation information over time.}
\label{Network_architecture}
\end{figure*}

Learning to see by moving is central to humans, starting very early during their perceptual and cognitive development\cite{gelman96}. Human observers are able to continuously improve the ability to understand their dynamic visual environment and accurately identify object categories, scene structures, actions or interactions. Supervision is necessary during early cognitive development but is practically impossible (and apparently unnecessary) at frame level. In contrast, current computer vision systems are not yet good at processing large amounts of complex video data, or at leveraging partial supervision within temporally coherent visual streams, in order to build increasingly more accurate semantic models of the environment. Successful semantic video segmentation models would be widely useful for indexing digital content, robotics, navigation or manufacturing. 

The problem of semantic image segmentation has received increasing attention lately, with some of the most successful methods being based on deep, fully trainable convolutional neural network (CNN) architectures. 
Data for training and refining single frame, static models is now quite diverse\cite{everingham2015pascal,lin2014microsoft}.
In contrast, fully trainable approaches to semantic video segmentation would face the difficulty of obtaining detailed annotations for the individual video frames, although datasets are emerging for the (unsupervised) video segmentation problem\cite{NB13}. Therefore, for now some of the existing, pioneering approaches to semantic video segmentation\cite{rehg2012,XuXiCoECCV2012,kundu2016feature} rely on single frame models with corresponding variables connected in time using random fields with higher-order potentials, and mostly pre-specified parameters. Fully trainable approaches to video are rare. The computational complexity of video processing further complicated matters. 

One possible practical approach to designing semantic video segmentation models in the long run can be to only label frames, sparsely, in video, as it was done for static datasets\cite{everingham2015pascal,lin2014microsoft}.  Then one should be able to leverage temporal dependencies in order to propagate information, then aggregate in order to decrease uncertainty during \emph{both} learning and inference. This would require a model that can integrate spatio-temporal warping across video frames. Approaches based on CNNs seem right -- one option could be to construct a fully trainable convolutional video network. 
Designing it naively could require e.g. matching edges in every possible motion direction as features. This would require a number of filters that is the number of filters for a single image CNN multiplied by possible motion directions. This holds for two frames. Extending it to even more frames would further amplify the issue. The problem here is how to choose a network architecture and how to connect different pixels temporally. Instead of building a spatio-temporal CNN directly, we will rely on existing single-frame CNNs but augment them with spatial transformer structures that implement warping along optical flow fields. These will be combined with adaptive recurrent units in order to learn to optimally fuse estimates from single (unlabeled) frames with temporal information from nearby ones, properly grated based on their uncertainty. The proposed model is differentiable and end-to-end trainable.

\section{Related work}

Our semantic video segmentation work relates to the different fields of semantic segmentation in images, as well as, more remotely, to (unsupervised) video segmentation and temporal modeling for action recognition. We will here only briefly review the vast literature with some attention towards formulations based on deep architectures which represent the foundation of our approach.

The breakthrough paper \cite{krizhevsky2012imagenet} introduced an architecture for image classification  where a deep convolutional network was trained on the Imagenet dataset. This has later been improved by \cite{simonyan2014very} who reduced the filter sizes and increased the depth substantially. Many semantic segmentation methods start with the network in \cite{simonyan2014very} and refine it for semantic segmentation. In \cite{he2015deep} residual connections are used making it possible to increase depth substantially. \cite{long2015fully} obtained semantic segmentations by turning a network for classification \cite{simonyan2014very} into a dense predictor by computing segmentations at different scales and then combining all predictions. The network was made fully convolutional. Another successful approach is to apply a dense conditional random field (CRF) \cite{krahenbuhl2011efficient} as a post-processing step on top of individual pixel or frame predictions. \cite{chen2014semantic} use a fully convolutional network to predict a segmentation and then apply the dense CRF as a post processing step. \cite{zheng2015conditional} realized that inference in dense CRFs can be formulated as a fix point iteration implementable as a recurrent neural network. The deep architecture our work is based on is \cite{yu2015multi} where max pooling layers are replaced with dilated convolutions. The network was extended by introducing a context module where convolutions with increasingly large dilation sizes are used.

Video segmentation has received significant attention starting from early methodologies based on temporal extensions to normalized cuts\cite{Shi:2000:NCI:351581.351611}, random field models and tracking\cite{rehg2012,lezama11}, motion segmentation\cite{DBLP:journals/pami/OchsMB14} or efficient hierarchical graph-based formulations\cite{grundman36247,XuXiCoECCV2012}. More recently proposal methods where multiple figure-ground estimates or multipart superpixel segmentations are generated at each time-step, then linked through time using optical flow\cite{li2013video,bais13iccv,Papazoglou:2013:FOS:2586117.2587271}, have become popular. \cite{kundu2016feature} used the dense CRF of \cite{krahenbuhl2011efficient} for semantic video segmentation by using pairwise potentials based on aligning the frames using optical flow.

Temporal modeling using deep architectures is also a timely topic in action recognition. One deep learning method \cite{simonyan2014two} uses two streams to predict actions. The authors use images in one stream and optical flow in the other stream, merged at a later stage in order to predict the final action label. This has also been implemented as a recurrent neural network in \cite{yue2015beyond} where the output of the two streams is sent into a recurrent neural network, where, at later stages the features are pooled to produce a final labelling. \cite{karpathy2014large} use frames at different timesteps and merge them as input to a neural network. \cite{sun2015human} attempt to make the problem of learning spatio-temporal features easier by using separable filters. Specifically, they first apply spatial filters and then temporal filters in order to avoid learning different temporal combinations for every spatial filter. \cite{ballas2015delving} learn spatio-temporal filters by stacking GRUs at different locations in a deep neural network and iterating through video frames. In a similar spirit \cite{li2016videolstm} use an LSTM and an attention mechanism to predict actions.

\section{Methodology}

\begin{figure*}
\begin{center}
\includegraphics[page=4,scale=0.45,trim=0cm 4.7cm 0cm 0cm]{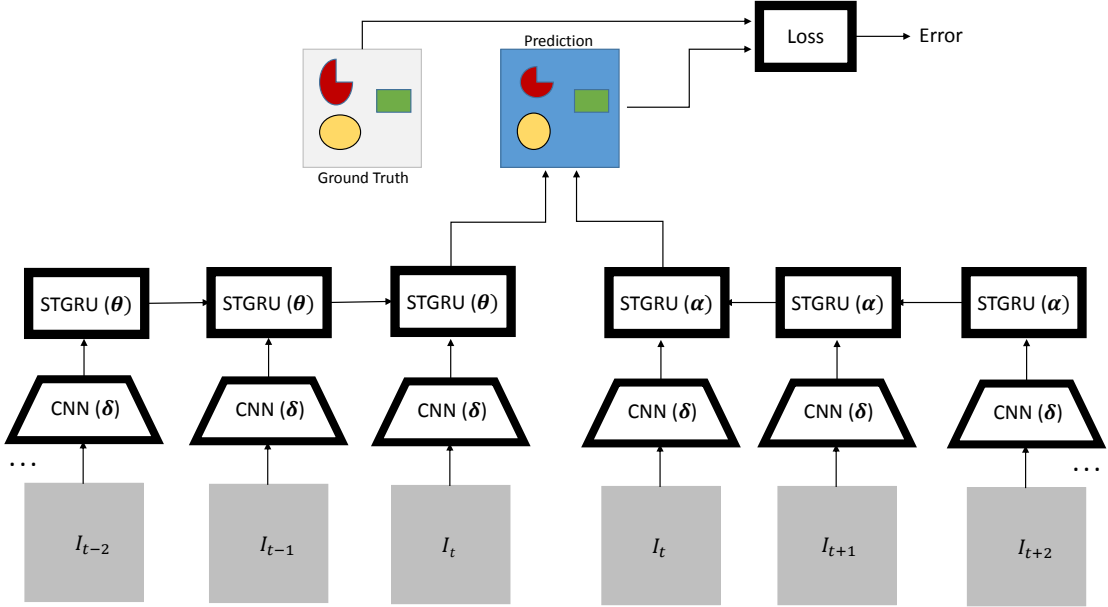}
\caption{Illustration of our temporal architecture entitled Gated Recurrent Flow Propagation (GRFP) based on Spatio-Temporal Transformer Gated Recurrent Units (STGRU), illustrated in fig.\ref{Network_architecture}. The model can integrate both forward only and forward-backward calculations, under separate recurrent units with different parameters $\theta$ and $\alpha$. Each of the forward and backward recurrent units have parameters that are tied across timesteps. The parameters of the semantic segmentation architecture are shared. The predictions from the forward model aggregated over frames $t-T,\ldots,t-1,t$ (in the above illustration $T=2$) and (when available and desirable) backward model aggregated over frames $t+T,\ldots t+1, t$ are fused at the central location $t$ in order to make a prediction that is then compared against the ground-truth available only at frame $t$ by means of a semantic segmentation loss function.}
\label{training_figure}
\end{center}
\end{figure*}

The visual illustration of how our semantic video segmentation model aggregates information in adjacent video frames is presented in fig. \ref{Network_architecture}. We start with a semantic segmentation at the previous time step, $h_{t-1}$ and warp it along the optical flow to align it with the segmentation at time $t$, by computing $w_t=\phi_{t-1,t}(h_{t-1})$ where $\phi$ is a mapping along the optical flow. This is fed as the hidden state to a gated recurrent unit (GRU) where the other input is the estimate $x_t$ computed by a single frame CNN for semantic segmentation. The information contained in $w_t$ and $x_t$ has significant redundancy, as one expects from nearby video frames, but in regions where it is hard to find the correct segmentation, or where significant motion occurs between frames, they might contain complementary information. The final segmentation $h_t$ combines the two segmentations by means of learnt GRU parameters and should include segments where either of the two are very confident. 

Our overall video architecture can operate over multiple timesteps both forward and backward with respect to the timestep $t$, say, where semantic estimates are obtained. In \emph{training}, the model has the desirable property that it can rely only on sparsely labeled video frames, but can take advantage of the temporal coherency in the unlabeled video neighborhoods centered at the ground-truth. Specifically, given an estimate of our static (per-image) semantic segmentation model at timestep $t$, as well as estimates prior and posterior to it, we can warp these using the confidence gated optical flow forward and backward in time (using the Spatio-Temporal Transformer Gated Recurrent Unit, STGRU, illustrated in fig. \ref{Network_architecture}) towards $t$ where ground truth information is available, then fuse estimates in order to obtain a prediction. The resulting model is conveniently differentiable. The loss signal will then be used to backpropagate information for training both the parameters of the gated recurrent units ($\theta$) and the parameters of the (per-frame) semantic segmentation network ($\delta$). The illustration of this mechanism is shown in fig. \ref{training_figure}. In \emph{testing} the network can operate either statically, per frame, or take advantage of video frames prior and (if available) posterior to the current processing timestep.

Given these intuitions we will now describe the main components of our model: the spatio-temporal transformer warping, the gated recurrent units, and the forward and backward implementations.

\subsection{Spatio-Temporal Transformer Warping}\label{sec:stnet}

We will use optical flow as input to warp the semantic segmentation estimates across successive frames. We will modify the spatial transformer network \cite{jaderberg2015spatial} for operation in the spatio-temporal video domain. Elements on a two-dimensional grid $x_{ij}$ will map to $y_{ij}$ according to
\begin{align}\label{eq:fi_map}
y_{ij} = \sum_{m,n} x_{mn} k(i + f_{ij}^y - m, j + f_{ij}^x - n) .
\end{align}
where $(f_{ij}^x, f_{ij}^y)$ is the optical flow at position $(i.j)$. We will use a bilinear interpolation kernel $k(x,y) = \max(0,1-|x|)\max(0, 1-|y|)$. The mapping is differentiable and we can backpropagate gradients from $y$ to both $x$ and $f$. The sum contains only $4$ non-zero terms when using the bilinear kernel, so it can be computed efficiently.

\subsection{Gated Recurrent Units for Semantic Video Segmentation}\label{sec:gru}

To connect the probability maps for semantic segmentation at different frames, $h_{t-1}$ and $h_t$, we will use a modified convolutional version of the Gated Recurrent Unit \cite{chung2014empirical}. In particular we will design a gating function based on the flow, so we only trust the semantic segmentation values warped from $h_{t-1}$ at location where the flow is certain. We also use gating to predict the new segmentation probabilities taking into account if either $h_{t-1}$ or $x_t$ have a high confidence for a certain class in some region of the image. Like an LSTM \cite{hochreiter1997long}, the GRU also have gating functions to reset hidden states and to control the influence of the input $x_t$ for the new hidden state $h_t$. The equations for a generic GRU with input $x_t$ and hidden state $h_t$ are
\begin{align}
& r_t = \sigma (W_{xr}x_t + W_{hr}h_{t-1}) \\
& \tilde{h}_t = \tanh (W_{xh} x_t + W_{hh} (r_t \odot h_{t-1})) \\
& z_t = \sigma (W_{xz} x_t + W_{hz} h_{t-1}) \\
& h_t = (1 - z_t) \odot h_{t-1} + z_t \odot \tilde{h}_t
\end{align}
where $r_t$ is a reset gate, $\tilde{h}_t$ is the candidate hidden state, $z_t$ is a gate controlling the weighing between the old hidden state $h_{t-1}$ and the candidate hidden state $\tilde{h}_t$, and $W$ are weight matrices for fully connected layers. We use $\odot$ to denote the element-wise product.

To adapt a generic GRU for semantic video segmentation, we first change all fully connected layers to convolutions. The hidden state $h_t$ and the input variable $x_t$ are no longer vectors but tensors of size $H \times W \times C$ where $H$ is the image height, $W$ is the image width and $C$ is the number of channels, corresponding to the different semantic classes. The input $x_t$ is normalized using softmax and $x_t(i,j,c)$ models the probability that the label is $c$ for pixel $(i,j)$. We let $\phi_{t-1,t}(x)$ denote the warping of a feature map $x$ from time $t-1$ to $t$, using optical flow given as additional input. The proposed adaptation of the GRU for semantic video segmentation is
\begin{align}
& w_t = \phi_{t-1,t}(h_{t-1}) \\
& r_t = 1 - \tanh(|W_{ir} *(I_t - \phi_{t-1,t}(I_{t-1})) + b_r|) \\
& \tilde{h}_t = W_{xh} *x_t + W_{hh} *(r_t \odot w_t) \\
& z_t = \sigma (W_{xz}* x_t + W_{hz}* w_t + b_z) \\
& h_t = \text{softmax}(\lambda (1 - z_t) \odot w_t + z_t \odot \tilde{h}_t) .
\end{align}
Instead of relying on a generic parametrization for the reset gate, we use a confidence measure for the flow by comparing the image $I_t$ with the warped image of $I_{t-1}$. We also discard $\tanh$ when computing $\tilde{h}_t$ and instead use softmax in order to normalize $h_t$. We multiply with $\lambda$ in order to compensate for a possibly different scaling of $\tilde{h}_t$ relative to the warped $h_{t-1}$ due to the convolutions with $W_{hh}$ and $W_{xh}$. Note that $h_{t-1}$ only enters when we compute the warping $w_t$ so we only use the warped $h_{t-1}$.

\subsection{Implementation}

For the static (per-frame) component of our model, we rely on a deep neural network pre-trained on the CityScapes dataset and fed as input to the gated recurrent units. We conducted experiments using both the Dilation architecture \cite{yu2015multi} and LRR \cite{lrr4x}.
The convolutions in the STGRU were all of size $7 \times 7$. We use the standard log-likelihood loss for semantic segmentation
\begin{align}
L(\theta) = -\sum_{i,j} \log p(y_{ij}=c_{ij} | I, \theta)
\end{align}
where $p(y_{ij}=c_{ij} | I, \theta)$ is the softmax normalized output of the STGRU. The recurrent network was optimized using Adam \cite{kingma2014adam} with $\beta_1=0.95$, $\beta_2=0.99$ and learning rate $2\cdot 10^{-5}$. Due to GPU memory constraints, the static computations had to be performed one frame at a time with only the final output saved in memory. When training the system end-to-end the intermediate activations for each frame had to be recomputed. We used standard gradient descent with momentum for refining the static network. The learning rate was $2\cdot 10^{-11}$ and momentum was $0.95$. Note that the loss was not normalized, hence the small learning rate. For flow we used Fullflow \cite{chen2016full} and we precomputed the flow between the frames closest to the labelled ones in the training set both forward and backward. For validation and testing we used DIS \cite{kroeger2016fast} which gave a higher accuracy. We used the version of DIS with the highest accuracy but the slowest runtime (0.5 Hz). We also conducted experiments where we used optical flow from a neural network \cite{fischer2015flownet} and refined the flow network jointly with the static network and the STGRU units. The images we trained on had output size 512x512, whereas in testing their size was increased to full resolution because intermediate computations do not have to be stored for backpropagation. 

\section{Experiments}

We perform an extensive evaluation on the challenging CityScapes and CamVid datasets, where video experiments remain nevertheless difficult to perform due to the large volume of computation. We evaluate under two different perspectives, reflecting the relevant, key aspects of our method. First we evaluate semantic video segmentation. We will compare our method with other methods for semantic segmentation and show that by using temporal information we can improve segmentation accuracy over a network where the predictions are per frame and unlabeled video data is not used. In the second evaluation we run our method in order to produce semantic segmentation for all frames in a longer video. We will then compare its temporal consistency against the baseline method where the predictions are performed per frame. We will show that our method gives a temporally much more consistent segmentation compared to the baseline.


\subsection{Semantic Video Segmentation}

\begin{table}[!htbp]
\begin{center}
\begin{tabular}{|c||c|c|}
\hline
Frames & mIoU classes & mIoU categories \\
\hline
\hline
1 & 0.687 & 0.863 \\
2 & 0.691 & 0.865 \\
3 & 0.693 & 0.865 \\
4 & 0.694 & 0.865 \\
5 & 0.694 & 0.865 \\
\hline 
\end{tabular}
\end{center}
\caption{Results on the validation set of CityScapes when using a different number of frames for inference. The model was trained using 5 frames. Notice that using more than one frame improves performance which however saturates beyond 4 frames.}
\label{Cityscapes_nbr_frames_table}
\end{table}

\begin{table}[!htbp]
\begin{center}
\small\begin{tabular}{|l||c|c|c|}
\hline
Class & Dilation10 & GRFP(5) & GRFP(1)\\
\hline
\hline
Road            & 0.972 & 0.973 & 0.969\\
Sidewalk        & 0.795 & 0.801 & 0.784\\
Building        & 0.904 & 0.905 & 0.904\\
Wall            & 0.449 & 0.510 & 0.479\\
Fence           & 0.524 & 0.530 & 0.523\\
Pole            & 0.551 & 0.552 & 0.557\\
Traffic light   & 0.567 & 0.573 & 0.565\\
Traffic sign    & 0.690 & 0.688 & 0.687\\
Vegetation      & 0.910 & 0.911 & 0.910\\
Terrain         & 0.587 & 0.598 & 0.588\\
Sky             & 0.926 & 0.930 & 0.927\\
Person          & 0.757 & 0.758 & 0.754\\
Rider           & 0.500 & 0.491 & 0.484\\
Car             & 0.922 & 0.925 & 0.921\\
Truck           & 0.562 & 0.575 & 0.560\\
Bus             & 0.726 & 0.741 & 0.729\\
Train           & 0.532 & 0.526 & 0.546\\
Motorcycle      & 0.462 & 0.486 & 0.457\\
Bicycle         & 0.701 & 0.708 & 0.708\\
\hline
Average         & 0.687 & 0.694 & 0.687 \\ \hline
\end{tabular}
\end{center}
\caption{Average IoU (mIoU) on the CityScapes validation set for the Dilation10 baseline, the best GRFP model using 5 frames, GRFP(5), and the refined Dilation10 net that the best GRFP learns, which is equivalent to GRFP(1).}
\label{detailed_scoreboard}
\end{table}

\begin{figure*}[!htbp]
\centering
\includegraphics[scale=0.314]{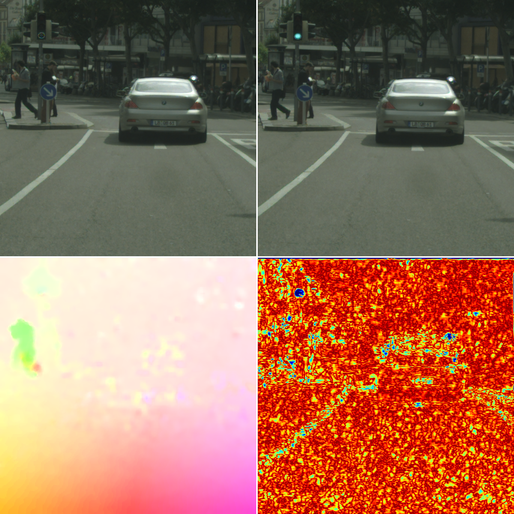}
\includegraphics[scale=0.314]{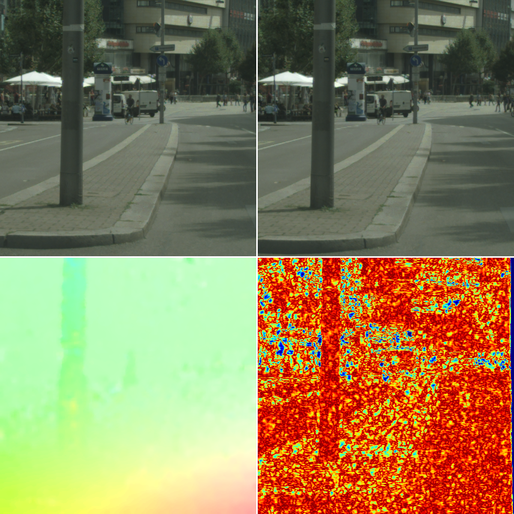}
\includegraphics[scale=0.314]{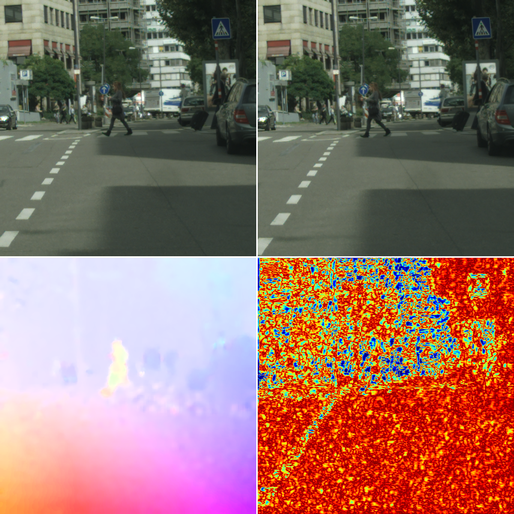}
\caption{Illustration of the flow gating as estimated by our Spatio-Temporal Transformer Gated Recurrent Unit. We show three pairs of consecutive frames, the flow and the its confidence as estimated by our STGRU model. Red regions indicate a confident flow estimate whereas blue regions are uncertain. Compare these to the errors of the optical flow method we use \cite{kroeger2016fast}, shown in fig.\ref{fig:flow-unc}. It is noticeable that our estimates integrate errors in the flow but exhibit additional uncertainty due to semantic labeling constraints in the STGRU model (\S\ref{sec:gru}).}
\label{flow_confidence}
\end{figure*}

\begin{figure}
\includegraphics[scale=0.15]{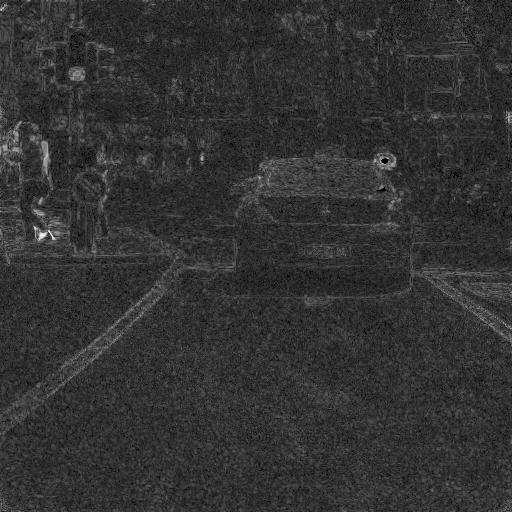}
\includegraphics[scale=0.15]{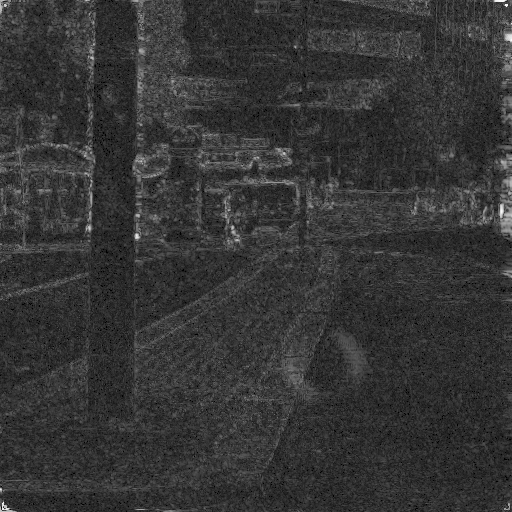}
\includegraphics[scale=0.15]{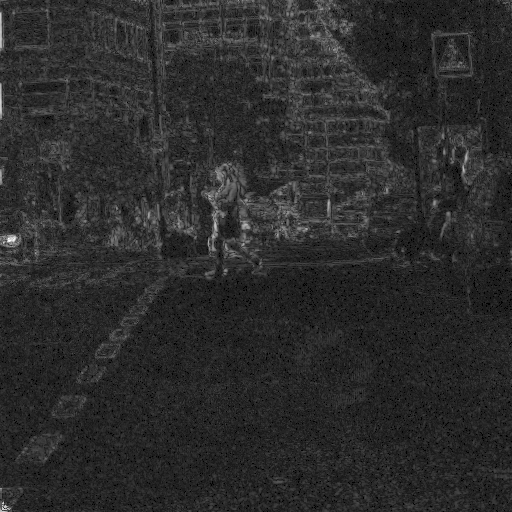}
\caption{Per-pixel errors in the optical flow estimation for three pairs of images shown in fig. \ref{flow_confidence}. Notice the increased uncertainty in fig. \ref{flow_confidence} as learnt by the STGRU model. }
\label{fig:flow-unc}
\end{figure}

\begin{table}[!htbp]
    \centering
    \begin{tabular}{|c||c|c|}
    \hline
        Method & mIoU class & mIoU category \\ \hline\hline
        PSP \cite{Zhao_2017_CVPR} & 81.2 & 91.2 \\ \hline
        GRFP + LRR-4x & 72.8 & 88.6 \\ \hline
        LRR-4x \cite{lrr4x} & 71.8 & 88.4 \\ 
        Adelaide\_context \cite{lin2015efficient} & 71.6 & 87.3 \\ 
        DeepLabv2-CRF \cite{chen2014semantic} & 70.4 & 86.4 \\
\hline
        GRFP + Dilation10 & 67.8 & 86.7 \\
        \hline
        Dilation10 \cite{yu2015multi} & 67.1 & 86.5 \\
        DPN \cite{liu2015semantic} & 66.8 & 86.0 \\
        FCN 8s \cite{long2015fully} & 65.3 & 85.7 \\ \hline
    \end{tabular}
    \caption{Results on the CityScapes test set for different published methods. We experimented with our methodology using both Dilation10 and LRR-4x as our baselines, and we are able to improve the mIoU accuracy with 0.7 and 1.0 percentage points respectively. All other methods only process a single frame and do not use video. Notice that our GRFP methodology proposed for video is applicable to most of the other semantic segmentation methods that predict each frame independently, and they can all benefit from potential performance improvements at no additional labeling cost.}
    \label{tab:test_set_results}
\end{table}

\begin{figure*}[!htbp]
    \centering
    \includegraphics[scale=0.308]{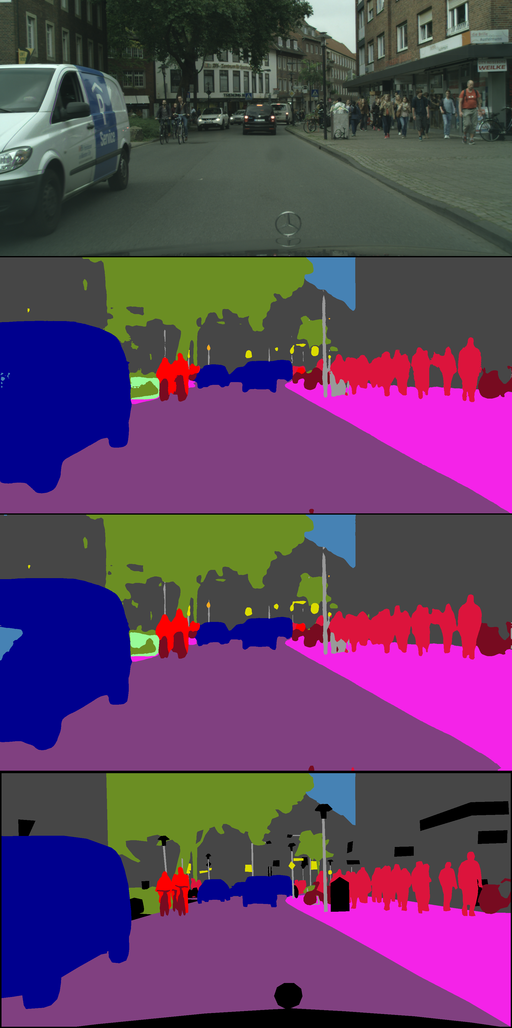}
    \includegraphics[scale=0.308]{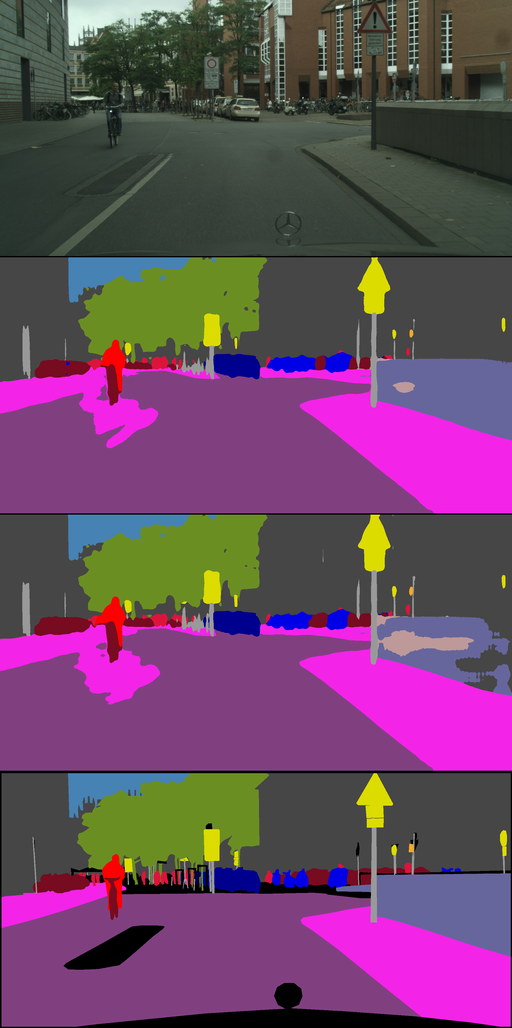}
    \includegraphics[scale=0.308]{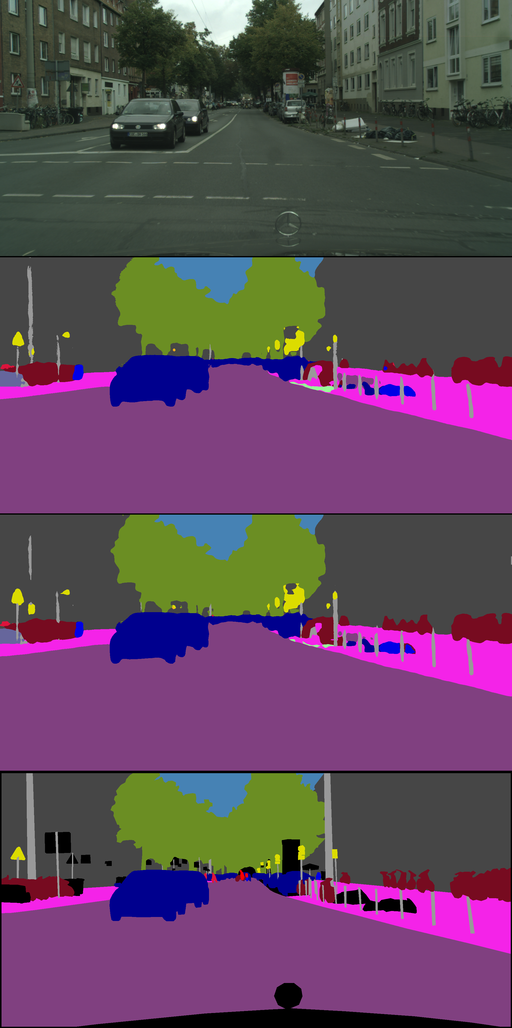}
    \caption{From top to bottom: the image, video segmentation by GRFP, static segmentation by Dilation10, and the ground truth. Notice the more accurate semantic segmentation of the car in the left example, the right wall in the middle example, and the left pole in the right example. For the first two examples the static method fails where the some object has a uniform surface over some spatial extent. The pole in the right image may be hard to estimate based on the current frame alone, but the inference problem becomes easier if earlier frames are considered, a property our GRFP model has.}
    \label{fig:ground_truth_comparison}
\end{figure*}

\begin{table*}[!htbp]
\begin{center}
\begin{tabular}{|c|c|c|c|c|c|c|c|c|c|}
\hline
     Method & Dilation10 & GRFP(5) & GRFP(1) & fwbw(a) & fwbw(b) & fwbw(c) & fwbw(d) & Flownet(e) & Flownet(f) \\ \hline
     mIoU & 0.687 & 0.694 & 0.687 & 0.694 & 0.696 & 0.695 & 0.694 & 0.692 & 0.690 \\ \hline
\end{tabular}
\end{center}
\caption{Results on the CityScapes validation set for various forward-backward models and for models using flownet and training end-to-end. See fig. \ref{training_figure} for how the parameters are defined. The first three models are described in more detail in Table \ref{detailed_scoreboard}. \textbf{(a)} A forward-backward model evaluated with $T=2$ using the same parameters for the backward STGRU as the best forward model GRFP(5) both in the forward direction and backward direction. We used $\theta$ and $\delta$ as for GRFP(5) and set $\alpha=\theta$. \textbf{(b)} as (a) but with $T=4$. \textbf{(c)} We used $\theta$ and $\delta$ from GRFP(5) but refined $\alpha$. We used $T=4$. \textbf{(d)} We refined all parameters $\theta$, $\alpha$ and $\delta$ and used $T=4$. (\textbf{e}) We used a setting identical to GRFP(5) but used flownet \cite{fischer2015flownet} to compute optical flow. (\textbf{f}) We used the setting in GRFP(5) and the network in \cite{fischer2015flownet} for optical flow and trained the Dilation network, the flow network and the recurrent network jointly.}
\label{extended_scoreboard}
\end{table*}

In CityScapes \cite{cordts2016cityscapes}, we are given sparsely annotated frames, and each labeled frame is the 20th frame in a 30 frame video snippet. The GRFP model we use for most experiments is a forward model trained using 5 frames (T=4 in fig. \ref{training_figure}). We apply the loss to the final frame. Notice however that due to computational considerations, while the STGRU unit parameters $\theta$ were trained based on propagating information from 5 frames, the unary network parameters $\delta$ were refined based on back-propagating gradient from the 2 STGRU units closest to the loss. Unless stated otherwise, we use the Dilation10 \cite{yu2015multi} network as our unary network.  

In Table \ref{Cityscapes_nbr_frames_table} we show the mean IoU (mIoU) over classes versus the number of frames used for inference. One can see that under the current representation, \emph{in inference}, not much gain is achieved by the forward model beyond propagating information from 4 frames. The results are presented in more detail in Table \ref{detailed_scoreboard} where we show the estimates produced by the pre-trained Dilation10 network and the per-frame Dilation network with parameters refined by our model, GRFP(1), as well as the results of our GRFP model operating over 5 frames GRFP(5). We also note that \cite{kundu2016feature} achieve a score of $0.703$ on the validation set but their method requires substantially more computation in inference (similar optical flow computation and unary prediction costs from the Dilation network, but differences between fully connected CRF belief propagation in \cite{kundu2016feature} versus convolutional predictions as one STGRU pass in our case). Notice also that while the average of our GRFP(1) is identical to the one of the pre-trained Dilation10, the individual class accuracies are different. It is apparent that most of our gains come from contributions due to temporal propagation and consistency reasoning in our STGRU models.

We also tried our model using LRR \cite{lrr4x} as backend to our model. As when we used the Dilation network, we got improved performance by using our video methodology compared to the static per-frame baseline. We show the detailed results on the validation set in Table \ref{tab:lrr_results} and test set scores are shown in Table \ref{tab:test_set_results}. Note that the accuracy is higher for all classes. For this experiment we used Flownet2 \cite{flownet2}.

In Table \ref{tab:test_set_results} we show semantic segmentation results on the CityScapes test set. We present the performance of a number of state of the art semantic segmentation models trained using Cityscapes but which do not use video. 
We used our GRFP methodology trained using both Dilation10 and LRR-4x as baseline models and in both cases we show  improved labelling accuracy. Notice also that our methodology can be used with any semantic segmentation method that processes each frame independently. Since we showed improvements using both baselines, we can predict that other single-frame methods can benefit from our proposed video methodology as well.

Figure \ref{fig:ground_truth_comparison} shows several illustrative situations where our proposed GRFP methodology outperforms the single frame baseline. In particular, our method is capable to more accurately segment the car, the right wall, and the left pole. In all cases it is apparent that inference for the current frame becomes easier when information is integrated over previous frames, as in GRFP.


\noindent {\bf Combining forward and backward models.} In Table \ref{extended_scoreboard} we show accuracy on the CityScapes validation set for various settings where we used the forward and backward models and averaged the predictions. This joint model was described in fig. \ref{training_figure}. We got the best result if we averaged predictions using $5$ frames going forward and adding the prediction going backward, that is, using $I_{t+4}, I_{t+3}, \hdots, I_t$. We also tried to combine the forward and backward predictions by stacking the two predictions and then having 3 layers where we learned to combine the predictions. We trained the new layers jointly with the forward and backward parameters $\theta$ and $\alpha$. This did not result in any improvement in accuracy.

\noindent {\bf Joint training including optical flow.} To make our model entirely end-to-end trainable we include optical flow estimation as a deep neural network component\cite{fischer2015flownet}. We conducted experiments where we used the flow from this network and refined everything, the STGRU parameters, the Dilation10 parameters and the flownet parameters jointly. Although the quality of the optical flow in \cite{fischer2015flownet} is significantly lower than the ones of other flow methods we used, training the model jointly still produced competitive results, see (e) and (f) in Table \ref{extended_scoreboard}. We note that the error signal passed to the flownet comes from a loss based on semantic segmentation. This is a very weak form of supervision to refine optical flow. Since the complete model is dependent on good optical flow and the STGRU units' performance decreases when using lower quality flow, the error signal to the flow network might be very noisy when starting the training, preventing efficient learning for the flownet. Note also that the STGRU units lose supervision in form of high quality optical flow. 


\begin{table}[!htbp]
\begin{center}
\small\begin{tabular}{|l|c|c|}
\hline
Class & LRR & LRR+GRFP\\
\hline
\hline
Road            & 0.977 & 0.978\\
Sidewalk        & 0.830 & 0.831\\
Building        & 0.912 & 0.914\\
Wall            & 0.496 & 0.514\\
Fence           & 0.568 & 0.581\\
Pole            & 0.607 & 0.611\\
Traffic light   & 0.656 & 0.675\\
Traffic sign    & 0.757 & 0.764\\
Vegetation      & 0.918 & 0.919\\
Terrain         & 0.622 & 0.626\\
Sky             & 0.940 & 0.942\\
Person          & 0.786 & 0.792\\
Rider           & 0.559 & 0.572\\
Car             & 0.934 & 0.937\\
Truck           & 0.606 & 0.624\\
Bus             & 0.762 & 0.786\\
Train           & 0.635 & 0.649\\
Motorcycle      & 0.487 & 0.532\\
Bicycle         & 0.730 & 0.741\\
\hline
Average         & 0.725 & 0.736 \\ \hline
\end{tabular}
\end{center}
\caption{Average IoU (mIoU) on the CityScapes validation set for the LRR baseline and our model GRFP based on LRR. By using our video methodology we can see labelling improvements for all classes.}
\label{tab:lrr_results}
\end{table}

\begin{figure*}[!htbp]
    \centering
    \includegraphics[scale=0.17]{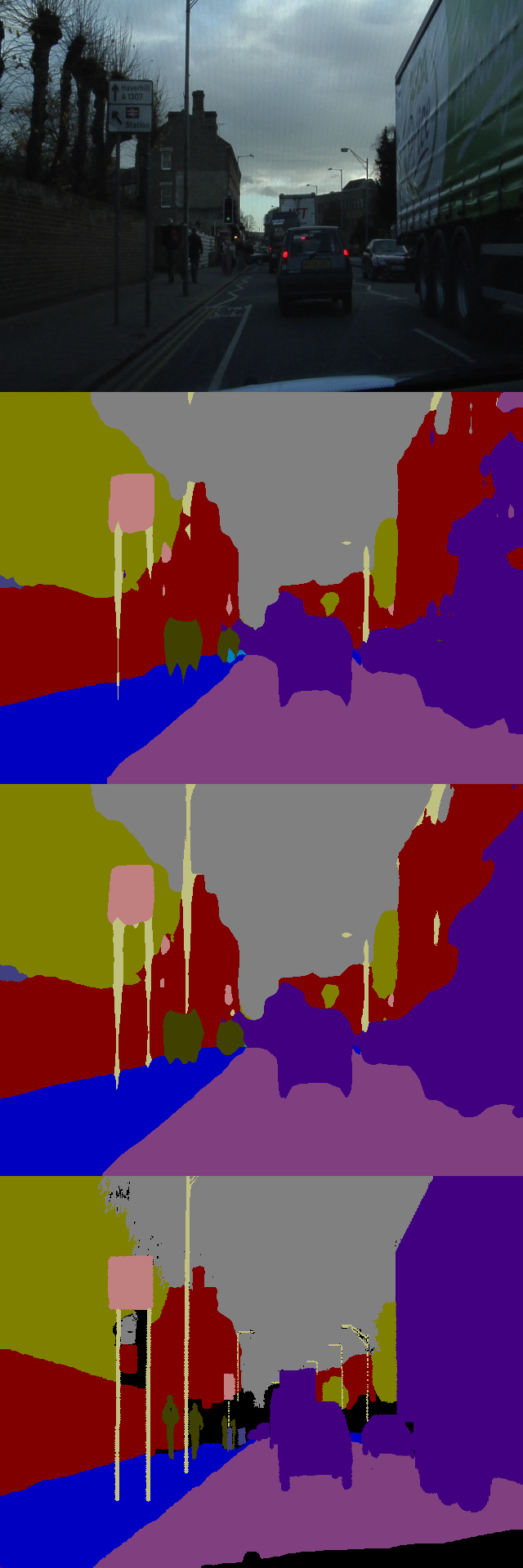}
    \includegraphics[scale=0.17]{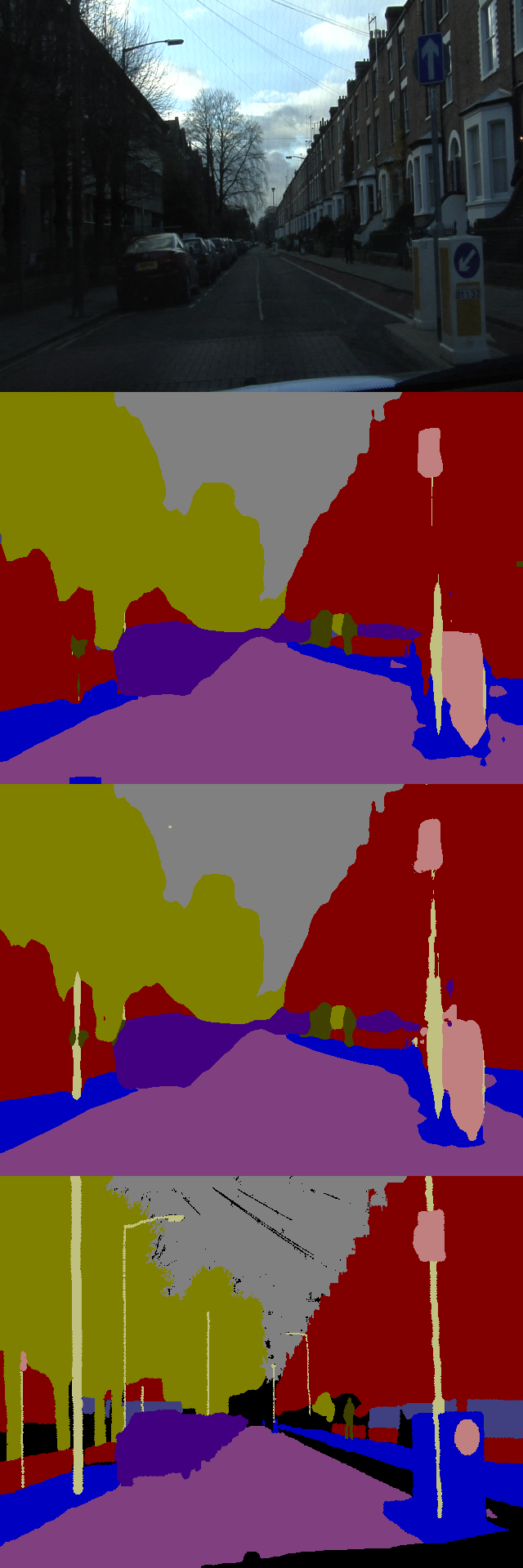}
    \includegraphics[scale=0.17]{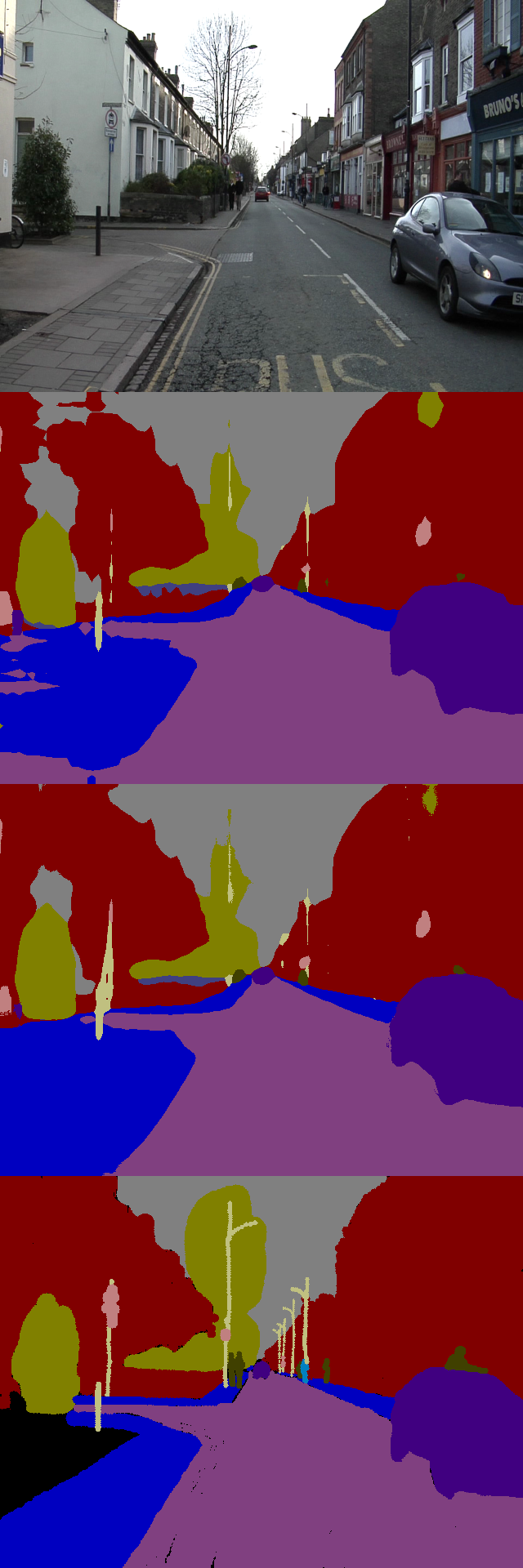}
    \includegraphics[scale=0.17]{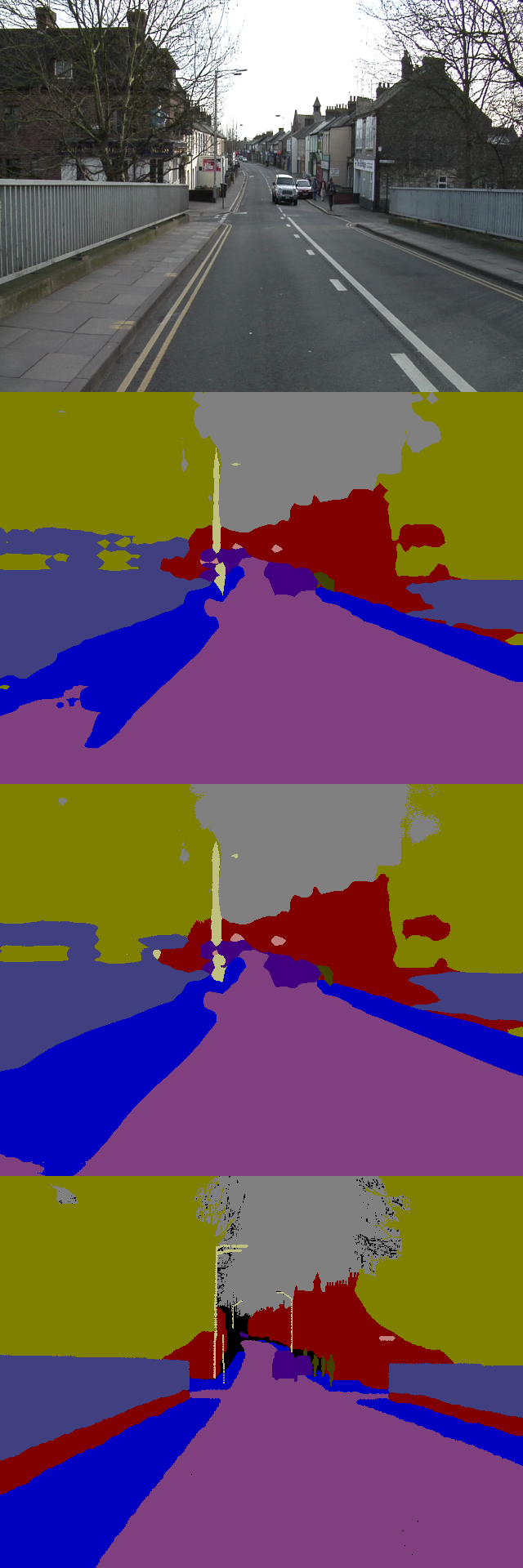}
    \caption{Qualitative examples from the CamVid test set. From top to bottom: the image, static segmentation by Dilation8, video segmentation by GRFP, and the ground truth. In the two examples to the left, notice that the poles are better segmented by our video method. In the two right images, the sidewalk is better segmented using our video methodology.}
    \label{fig:camvid}
\end{figure*}

\noindent {\bf CamVid} To show that our method is not limited to CityScapes we also provide additional experiments on the CamVid dataset \cite{brostow2009semantic}. This dataset consists of 4 video sequences that are annotated at 1 Hz. In total there are 367 training images, 100 validation images and 233 test images. We use the Dilation network \cite{yu2015multi} as our baseline and we use our GRFP methodology to improve the performance by using video and not processing each frame independently. The results can be seen in Table \ref{tab:camvid_results}. We can see that we are able to improve the segmentation accuracy by using additional video frames as input, and our accuracy results are on par with the state-of-the-art method \cite{kundu2016feature}. We were not able to perfectly replicate the numbers reported in \cite{yu2015multi}, but the mean IoU is the same and the class differences are marginal. We show several qualitative examples in fig. \ref{fig:camvid}.

\begin{table*}[!htbp]
    \centering
    \begin{tabular}{|c||c|c|c|c|c|c|c|c|c|c|c||c|} \hline
        & \rotatebox{90}{Building} & \rotatebox{90}{Tree} & \rotatebox{90}{Sky} & \rotatebox{90}{Car} & \rotatebox{90}{Sign} & \rotatebox{90}{Road} & \rotatebox{90}{Pedestrian} & \rotatebox{90}{Fence}  & \rotatebox{90}{Pole}  & \rotatebox{90}{Sidewalk}  & \rotatebox{90}{Bicycle} & \rotatebox{90}{mean IoU} \\ \hline
        SegNet \cite{badrinarayanan2015segnet} & 68.7 & 52.0 & 87.0 & 58.5 & 13.4 & 86.2 & 25.3 & 17.9 & 16.0 & 60.5 & 24.8 & 46.4 \\ \hline
        DeepLab-LFOV \cite{chen2014semantic} & 81.5 & 74.6 & 89.0 & 82.2 & 42.3 & 92.2 & 48.4 & 27.2 & 14.3 & 75.4 & 50.1 & 61.6 \\ \hline
        Dilation8 \cite{yu2015multi} & 83.0 & 76.3 & 90.3 & 84.1 & 47.9 & 92.7 & 56.1 & 36.8 & 20.2 & 76.4 & 54.1 & 65.3 \\ \hline
        GRFP & 83.1 & 76.6 & 90.3 & 84.6 & 48.7 & 93.3 & 56.8 & 37.1 & 24.3 & 77.9 & 54.7 & 66.1 \\ \hline
        FSO \cite{kundu2016feature} & 84.0 & 77.2 & 91.3 & 85.6 & 49.9 & 92.5 & 59.1 & 37.6 & 16.9 & 76.0 & 57.2 & 66.1 \\ \hline
    \end{tabular}
    \caption{Results on the test set of CamVid. Note that out method GRFP gets a higher score than Dilation8 which it is based on.}
    \label{tab:camvid_results}
\end{table*}

\subsection{Temporal Consistency}

\begin{figure*}[!htbp]
    \includegraphics[scale=0.4]{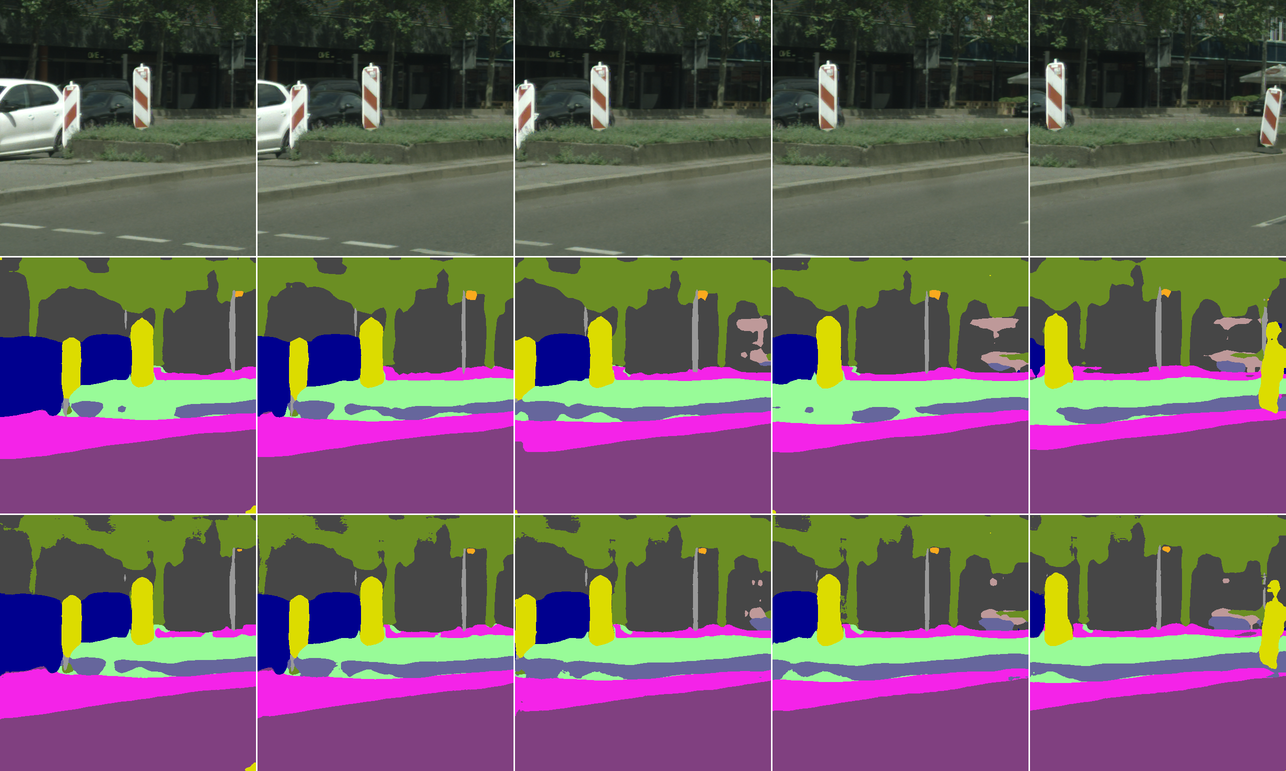}
    \includegraphics[scale=0.4]{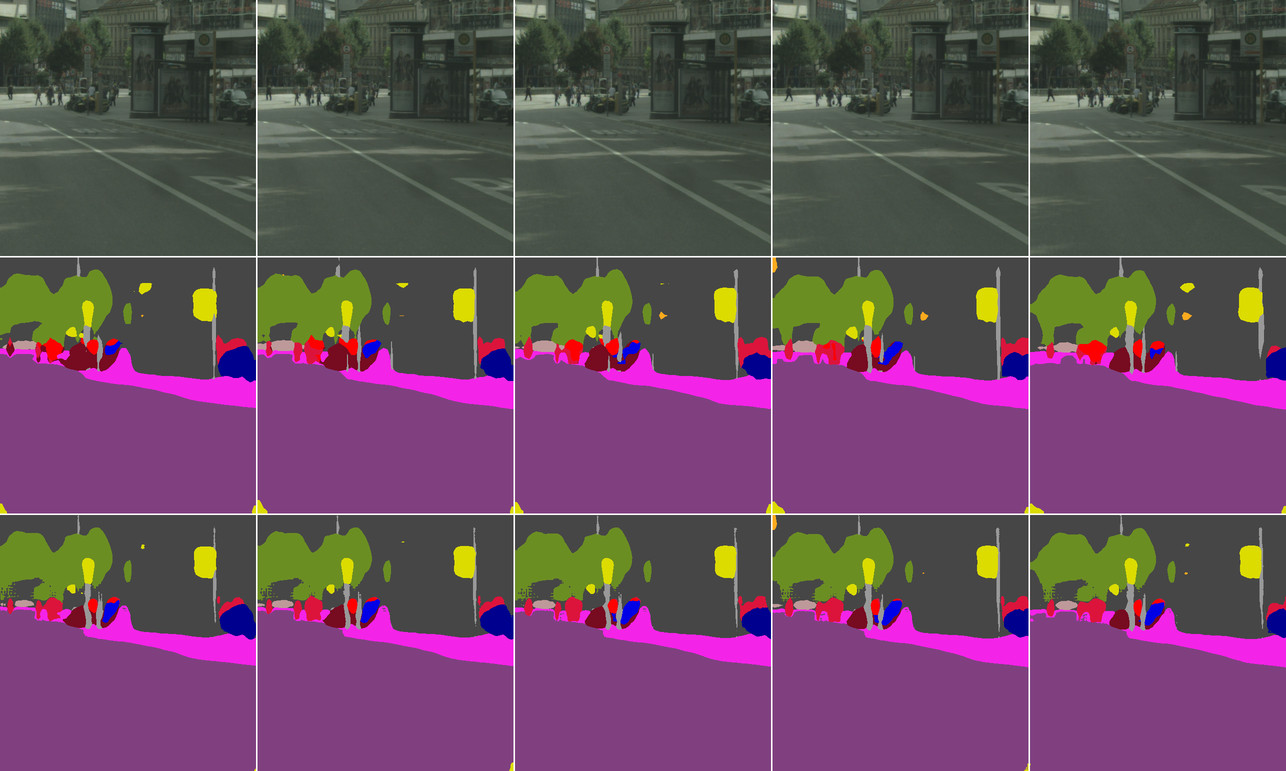}
    \caption{Examples of semantic segmentations in consecutive video frames from CityScapes. From top to bottom we have the frames, the static segmentation by Dilation10 and the video segmentation by GRFP. In the first example we have a more consistent estimation of vegetation over time and in the second example we see that the GRFP suppresses some spurious objects in the middle of the image that appear in a few of the frames for the Dilation network.}
    \label{video_examples}
\end{figure*}

As in \cite{kundu2016feature}, we evaluate the temporal consistency of our semantic video segmentation method by computing trajectories in the video using \cite{sundaram2010dense} and calculating for how many of the trajectories the labelling is the same in all frames. The results are shown in Table \ref{tab:temporal_results}. We use the demo videos provided in the CityScapes dataset, that are 600, 1100 and 1200 frames long, respectively. Due to computational considerations, we only used the middle 512x512 crop of the larger CityScapes images. The results can be seen in Table \ref{tab:temporal_results} where improvements are achieved for all videos at an average of about 4 percentage points. Qualitatively, there is a lot less flickering and noise when we are using the proposed GRFP semantic video segmentation methodology compared to models that rely on single-frame estimates. Semantic segmentations for consecutive frames in a video for both GRFP and the single-frame Dilation10 network are shown in fig. \ref{video_examples}.

\begin{table}[!htbp]
    \centering
    \begin{tabular}{|c||c|c|c|} \hline
        Video & Dilation10 & GRFP & FSO \cite{kundu2016feature} \\ \hline\hline
        stuttgart\_00 & 79.18 \% & 84.29 \% & 91.31 \% \\ 
        stuttgart\_01 & 86.13 \% & 88.87 \% & 93.32 \%  \\ 
        stuttgart\_02 & 76.77 \% & 81.96 \% & 89.01 \%  \\ \hline
        Average & 80.69 \% & 85.04 \% & 91.21 \% \\ \hline
    \end{tabular} 
    \caption{Temporal consistency for the different demo videos in the CityScapes dataset. We note that our GRFP semantic video segmentation method achieves a more consistent segmentation than the single frame baseline.}
    \label{tab:temporal_results}
\end{table}

\subsection{Timing}

We report the timing of the different components of our method using a Titan X GPU. For the flow computations we report the time using either Flownet1 \cite{fischer2015flownet} or Flownet2 \cite{flownet2}. The numbers in Table \ref{tab:timing} show the timing per frame of the three main components of our framework: the static prediction, the flow computation and the STGRU computations. We report the time to process (testing, not training) one frame in a video with resolution 512x512. With our methodology we can obtain an improved temporal consistency and labelling accuracy with an additional runtime of 335 ms per frame with Flownet2 and 75 ms with Flownet1.

\begin{table}[]
    \centering
    \begin{tabular}{|c|c|c|} \hline
         & Dilation10 & LRR \\ \hline
        Segmentation module & 350 ms & 200 ms \\ 
        Flownet2/Flownet1 & 300/40 ms & 300/40 ms \\ 
        STGRU & 35 ms & 35 ms \\ \hline
    \end{tabular}
    \caption{Timing of the different components of out GRFP methodology using a Titan X GPU. We are able to show improved segmentation accuracy and temporal consistency by incurring an additional runtime of 75 ms per frame if we use Flownet1 or 335 ms per frame if we use Flownet2.}
    \label{tab:timing}
\end{table}

\section{Conclusions}
We have presented a deep, end-to-end trainable methodology for semantic video segmentation (including the capability of jointly refining the recognition, optical flow and temporal propagation modules), that is capable of taking advantage of the information present in unlabeled frames in order to improve estimates. Our model combines a convolutional architecture and a spatio-temporal transformer recurrent layer that learns to temporally propagate labeling information by means of optical flow, adaptively gated based on its locally estimated uncertainty. Our extensive experiments on the challenging CityScapes and CamVid datasets, and for multiple deep semantic models, indicate that our resulting model can successfully propagate information from labeled video frames towards nearby unlabeled ones in order to improve both the semantic video segmentation accuracy and the consistency of its temporal labeling, at no additional annotation cost and with little supplementary computation. \\

\noindent{\bf Acknowledgements:} This work was funded in part by the European Research Council, ERC, Consolidator Grant SEED.

{\small
\bibliographystyle{ieee}
\bibliography{egbib}
}

\end{document}